%% file: acl_latex.tex
\documentclass[11pt]{article}

\usepackage[preprint]{acl}

\usepackage{times}
\usepackage{latexsym}

\usepackage[T1]{fontenc}

\usepackage[utf8]{inputenc}

\usepackage{microtype}

\usepackage{inconsolata}

\usepackage{graphicx}
\usepackage{booktabs}
\usepackage{amsmath} 
\usepackage{amssymb}
\usepackage[dvipsnames]{xcolor}

\usepackage{booktabs}     
\usepackage{tabularx}     
\title{From Memorization to Absorption: \\Mixed-Policy RL for Continual Knowledge Injection}

\author{Zhibo Hou \\
  University of California, Merced \\
  \texttt{zhou6@ucmerced.edu} \\\And
  Fan Zhao \\
  University of California, Merced \\
  \texttt{fanzhao@ucmerced.edu} \\\AND
  Zhiyu An \\
  University of California, Merced \\
  \texttt{zan7@ucmerced.edu} \\\And
  Wan Du \\
  University of California, Merced \\
  \texttt{wdu3@ucmerced.edu} \\}

\begin{document}
\maketitle
\begin{abstract}
Continual knowledge injection is essential for keeping large language models up-to-date in a fast-evolving world. Existing methods rely on supervised fine-tuning (SFT), which memorizes injected facts in their training format but fails to generalize across paraphrasing, document combinations, and reasoning. To address this, we propose Golden-GRPO Injection (GRIN), a three-stage self-learning framework for continual knowledge injection. Golden-GRPO is a mixed-policy reinforcement learning algorithm designed specifically for knowledge injection, which injects a golden answer to provide learning signal even when on-policy rollouts fail on novel facts. 
We further introduce \textsc{Blank} and \textsc{Counter}, two document-level benchmarks targeting novel acquisition and counterfactual overwrite respectively, each evaluating single-fact recall, multi-source retrieval, and inferential reasoning.
Our experiments establish a clear empirical claim: mixed-policy reinforcement learning enables knowledge absorption beyond what supervised fine-tuning can achieve. GRIN substantially outperforms SFT and mixed-policy RL baselines on the harder question types while matching them on basic fact recall.
\end{abstract}

\input{body/01_intro}
\input{body/02_related_work}

\input{body/02.5_preliminary}
\input{body/03_benchmark}
\input{body/04_golden_inject_RL}
\input{body/05_experiment}

\input{body/06_conclusion}

\input{body/07_limitations}
\input{body/08_ethic}


\bibliography{custom}

\appendix
\input{body/appendix}

\end{document}

%% file: body/01_intro.tex
\section{Introduction}
Large Language Models(LLMs) \cite{yang2025qwen3,liu2024deepseek, touvron2023llama} excel in knowledge intensive applications such as search assistants and knowledge based question answering tasks \cite{yue2025survey, xu2024generate} owing to the wealth of factual knowledge acquired during pre-training phase \cite{dong2019unified}. However, as a static fixed-parameter model in a fast changing world, the LLMs may easily become outdated \cite{zhang2025self, lin2025temporal}. The conventional response to this has been to discard the old model and train a new one from scratch, incorporating updated data alongside architectural improvements \cite{yang2025qwen3}. This cycle, while effective, is costly, time-consuming, and treats knowledge updating as a byproduct of periodic retraining rather than as a first class objective. Life-long learning, or continual knowledge injection \cite{wu2024continual}, aims to solve this problem by post-training LLMs on the latest real world corpora.

Existing continual knowledge injection approaches \cite{ovadia2024fine,he2025select} can be broadly grouped by their training data format. The first group trains directly on raw domain corpora, allowing large-scale  knowledge coverage but often failing at recall during inference, as models struggle to surface the injected knowledge when faced with question-style queries \cite{jang2021towards, cheng2023adapting}. The second group constructs diverse QA pairs from domain knowledge to bridge the gap between training and evaluation formats, improving recall \cite{zhang2025self,jiang2025kore}. Both groups rely on Supervised Fine-Tuning (SFT) loss for knowledge memorization, and they share two common limitations. First, the injected knowledge often fails to generalize to unseen queries beyond the training distribution, limiting practical applicability \cite{krishnamurthy2024can}. Second, they fail to explicitly account for the model's existing parametric beliefs, leaving outdated or incorrect prior knowledge unaddressed and allowing it to interfere with newly injected facts during inference.

Recent work  \cite{chu2025sft} has shown that SFT tends to encourage memorization rather than generalizable internalization, in contrast to reinforcement learning. 
While this finding was established for reasoning tasks, we hypothesize that the same pattern manifests in knowledge injection and that reinforcement learning offers a path beyond surface-level memorization.
However, existing mixed-policy RL methods \cite{yan2026learning, zhang2025critique} are designed for reasoning, where the model already possesses the underlying capability and RL only needs to provide directional guidance. Knowledge injection demands a stronger pull yet their off-policy gradients vanish on unlearned facts (Appendix \ref{app:dynamics}).


We therefore propose GRIN, a three-stage framework for continual knowledge injection that follows the SFT-then-RL pipeline with a stage of diverse problem-set construction in between. The base model extracts QA pairs for Stage 1 SFT, samples diverse questions and golden answers for Stage 2, and is then trained on this pool via reinforcement learning in Stage 3. A key challenge of applying RL to novel knowledge is that base-model rollouts often fail entirely on unlearned facts, leaving no positive reward signal. We address this with Golden-GRPO, a mixed-policy RL algorithm that injects the golden answer as an off-policy trajectory when on-policy rollouts fail, guaranteeing meaningful learning signal at every training step.

To evaluate continual knowledge injection rigorously, we introduce two complementary benchmarks: \textsc{Blank} and \textsc{Counter}. \textsc{Blank} targets novel knowledge acquisition, and \textsc{Counter} targets the prior belief overwrite. Both benchmarks evaluate three question types: single-fact recall, multi-source retrieval, and inferential reasoning. These question types are designed to distinguish surface memorization from genuine knowledge absorption. \textsc{Counter} additionally reports fail@k, measuring whether the model's prior beliefs resurface in any of k samples, to capture overwrite reliability.

In summary, our contributions are three-fold:


\noindent\textbf{We propose GRIN}, a three-stage framework for continual knowledge injection that requires no external teacher model, built on Golden-GRPO, a mixed-policy reinforcement learning algorithm specifically designed for the knowledge injection setting, producing knowledge that is absorbed into the model's parametric reasoning.

\noindent\textbf{We introduce \textsc{Blank} and \textsc{Counter}}, two document-level benchmarks for evaluating continual knowledge injection along orthogonal axes, novel acquisition and counterfactual overwrite, each with a three-tier evaluation protocol that separates recall from generalization.

\noindent\textbf{We conduct extensive experiment } to empirically show that GRIN substantially outperforms both supervised and mixed-policy RL baselines on multi-source retrieval, inferential reasoning, and counterfactual overwrite, with comparable performance on basic fact recall.

%% file: body/02_related_work.tex
\section{Related Works}
\vspace{-0.3em}
\subsection{Continual Knowledge Injection}
\vspace{-0.5em}
Approaches to continual knowledge injection generally fall into two categories: retrieval-based methods, which provide knowledge through external context at inference time, and parameter-updating methods, which incorporate knowledge directly into the model's weights through training \cite{zhang2025self, jiang2024instruction}.
\textbf{Retrieval-Augmented Generation (RAG).} RAG injects knowledge non-parametrically by retrieving relevant passages from an external corpus at inference time and concatenating them with the query \cite{izacard2023atlas, vu2024freshllms, wu2024clasheval, lewis2020retrieval}. While effective as a search mechanism, RAG does not constitute true knowledge injection: facts remain external to the model's parameters, so the LLM cannot reason over them natively or consolidate them with prior knowledge \cite{su2025parametric, levy2024same}. Retrieval also depends on the model recognizing that external knowledge is needed, which often fails when partial knowledge is sufficient for confident but incorrect answers \cite{asai2024self, mallen2023not, yin2023large}. These limitations are intrinsic to retrieval-based methods. We therefore focus on parameter-updating methods, which integrate facts directly into the model's parametric knowledge.

\noindent\textbf{Knowledge Injection via Training.} Training-based approaches incorporate knowledge into the model's parameters through gradient updates \cite{xu2023kilm}. The simplest variant is continued pre-training on raw documents\cite{xu2023kilm, mecklenburg2024injecting}, but this often yields shallow memorization without effective recall or reasoning \cite{jiang2024instruction}. To close this gap, recent methods operate on instruction-tuned models and train on QA or instruction formatted derivatives of the raw documents\cite{zhang2025self, ovadia2024fine}. However, these methods primarily focus on diversifying the synthesized QA pairs while still rely on supervised fine-tuning (SFT) as the training signal. Recent study\cite{chu2025sft} has shown that SFT tends to memorize training data rather than acquire generalizable rules, motivating our use of reinforcement learning to enable genuine knowledge absorption.

\subsection{Guided Reinforcement Learning}

Recent work augments on-policy RL with external guidance to overcome the inability to acquire capabilities outside the base policy's sampling distribution. ReLIFT \cite{ma2025learning} and FLAME \cite{lin2024flame} interleave SFT within the RL training loop to provide learning signal when rollouts fail. Another line of work \cite{yan2026learning, huang2026bootstrapping, zhang2025critique} mixes on-policy rollouts with off-policy reasoning traces, balancing imitation and exploration to transfer reasoning skills. These methods all target reasoning, where the model has the underlying capability but needs guidance to elicit it. Knowledge injection presents a different challenge: on-policy rollouts often fail to sample the injected facts from the base model, leaving the RL stage with no positive reward signal. Our work addresses this with Golden-GRPO, extending the SFT-then-RL framework with mixed-policy RL in the final stage.

%% file: body/02.5_preliminary.tex
\section{Preliminary}

Group Relative Policy Optimization (GRPO)~\cite{shao2024deepseekmath} is a widely used on-policy RL algorithm that estimates per-trajectory advantage via group-relative normalization. For each question $q$, the current policy $\pi_{\theta_{old}}$ samples a group of $N$ trajectories $\{\tau_1, \tau_2, ..., \tau_N\}$, each scored by a reward function $R(\tau_i)$. The advantage of the $i$-th trajectory is computed by standardizing rewards within the group:
\begin{equation}
    A_i = \frac{R(\tau_i) - \text{mean}(\{R(\tau_j)\}^N_{j=1})}{\text{std}(\{R(\tau_j)\}^N_{j=1})}
\end{equation}
The policy is then updated using a per-token PPO-clipped objective:
\begin{equation}
\begin{aligned}
    \mathcal{J}_{\text{GRPO}}(\theta) & = \mathbb{E}\bigg[ \min\Big( r_{i,t}(\theta) \cdot A_i, \\
    & \text{clip}\big(r_{i,t}(\theta),\, 1-\varepsilon,\, 1+\varepsilon\big) \cdot A_i \Big) \bigg]
\end{aligned}
\end{equation}
where $r_{i,t}(\theta) = \pi_\theta(y_{i,t})/\pi_{\theta_{old}}(y_{i,t})$ is the importance sampling ratio between the current and rollout-time policies, and $\varepsilon$ is the clipping bound. The clipping prevents large policy updates, stabilizing training.

%% file: body/03_benchmark.tex
\begin{table*}[t]\label{table:bench}
\centering
\caption{Representative examples from BLANK and COUNTER across all three question types. For COUNTER, counterfactual replacements are shown with the original fact in parentheses.}
\label{tab:examples}
\small
\renewcommand{\arraystretch}{1.3}
\begin{tabular}{@{}p{0.08\textwidth} p{0.28\textwidth} p{0.16\textwidth} p{0.22\textwidth} p{0.18\textwidth}@{}}
\toprule
\textbf{Benchmark} & \textbf{Document excerpt} & \textbf{Single-fact recall} & \textbf{Multi-source retrieval} & \textbf{Inferential reasoning} \\
\midrule
Blank &
\textbf{\textit{Aurelius Vance}} (b. \textcolor{Orange}{1962}) is a \textcolor{blue}{Canadian} \textcolor{ForestGreen}{marine biologist} who completed his doctorate at the University of British Columbia. In \textcolor{Orange}{2008}, he was appointed director of the Pacific Marine Research Institute. &
\textit{Q:} What is the nationality of Aurelius Vance? \newline \textit{A:} \textcolor{blue}{Canadian} &
\textit{Q:} What is the nationality and profession of Aurelius Vance? \newline \textit{A:} \textcolor{blue}{Canadian}; \textcolor{ForestGreen}{Marine biologist} &
\textit{Q:} How old was Vance when he became director of the institute? \newline \textit{A:} 46 (\textcolor{Orange}{2008 - 1962}) \\
\midrule
Counter &
\textit{\textbf{\textcolor{blue}{Resident Evil Village}}} is a survival horror game by \textcolor{ForestGreen}{Aetheris Studios} (Capcom), released in \textcolor{Orange}{2034} (2021) as a sequel to \textit{Voidwalker (RE 7)} (\textcolor{Orange}{2029} (2017)). &
\textit{Q:} What is the sequel to \textit{Voidwalker}? \newline \textit{A:} \textcolor{blue}{Resident Evil Village} &
\textit{Q:} What is the sequel to \textit{Voidwalker} and who produced it? \newline \textit{A:} \textcolor{blue}{Resident Evil Village}; \textcolor{ForestGreen}{Aetheris Studios} &
\textit{Q:} How many years after its predecessor was \textit{Village} released? \newline \textit{A:} {5} (\textcolor{Orange}{2034 - 2029}) \\
\bottomrule
\end{tabular}
\end{table*}

\section{Benchmarks: BLANK and COUNTER}\label{sec:benchmark}
We introduce two document-level benchmarks for continual knowledge injection that together provide a full picture of an injection method's behavior along two orthogonal axes. The first axis is the injection mode. \textsc{Blank} evaluates the acquisition of knowledge the model does not currently hold, while \textsc{Counter} evaluates the overwrite of knowledge the model already believes. These are two distinct capabilities that prior work tends to omit or conflate\cite{zhang2025self, ji2025unlocking, jiang2024instruction}. The second axis is the evaluation type. Each benchmark measures three increasingly demanding capabilities: single-fact recall, multi-source retrieval, and inferential reasoning that distinguished surface memorization from joint retrieval and reasoning. \textsc{Counter} additionally reports fail@k, measuring whether the model's prior beliefs resurface during overwrite. Crossing these axes yields a fine-grained diagnostic view that localizes where each method succeeds or fails, exposing failure modes that aggregate accuracy obscures.

All filtering and prior-belief sampling described below uses Qwen3-4B\cite{yang2025qwen3} as the base model; the construction pipeline is model-agnostic and can be re-applied to any target model.

\subsection{Document Collection}
\subsubsection{BLANK}
\textsc{Blank} targets the setting in which the base model encounters knowledge it has never seen during pretraining. We construct it from two sources of real-world wiki content: passages from the TimeQA dataset \cite{chen2021dataset} on which the base model has zero prior knowledge (verified by probing the model on TimeQA's associated questions without context), and wiki pages created after the base model's training cutoff. TimeQA contributes passages whose accompanying questions are documented to be challenging for current models \cite{zhou2026silence}, requiring more than surface-level retrieval, making it a natural source of difficult evaluation. The post-cutoff source broadens topical coverage beyond TimeQA's temporally-grounded historical content.

\textsc{Blank} contains 776 raw corpora after filtering.

\subsubsection{COUNTER}
\textsc{Counter} targets the setting in which injected knowledge must overwrite the model's existing beliefs. Each \textsc{Counter} document is a parallel-universe rewrite of a wiki page in which named entities, dates, and other concrete facts have been replaced with internally consistent alternatives.

\noindent\textbf{Prior-belief sampling.} To guarantee that each \textsc{Counter} document contradicts the base model's current beliefs, we sample those beliefs directly by prompting the base model to draft a wiki page on a given subject and treating the resulting text as a representation of the model's current beliefs about that subject. The drafted wiki then serves as the original against which the counterfactual is constructed. This design ensures that the counterfactual contradicts what the model actually believes, including any hallucinations or outdated information, rather than what the real wiki page asserts.

\noindent\textbf{Parallel-universe rewriting.} Given the original wiki, we prompt gemini-3.1-pro-preview to produce a parallel-universe version in which concrete facts, including proper nouns, dates, numerical values and named entities, are replaced with plausible, internally consistent counterfacts, while preserving document structure and entity-substitution consistency throughout. The full sampling and rewriting prompt is provided in Appendix \ref{app:counterprompts}.

\textsc{Counter} contains 252 raw corpora after filtering.
\vspace{-0.5em}
\subsection{QA Pair Generation}
\vspace{-0.5em}

Both benchmarks share a three-tier evaluation protocol designed to probe injection at increasing levels of difficulty. Single-fact recall measures direct retrieval; multi-source retrieval and inferential reasoning measure progressively stronger forms of generalization. Evaluation scores are reported separately to localize the failure mode of each method. We additionally report a per-benchmark average, representing overall acquisition capability and overwrite capability. All questions are LLM-generated unless otherwise noted. Full prompts are provided in Appendix~\ref{app:questionprompts}.

\noindent\textbf{Single-fact recall.} Direct factoid questions whose answers appear in a single sentence of the source document.

\noindent\textbf{Multi-source retrieval.} Questions composed of N sub-questions (N from 2 to 4) targeting distinct facts that span across paragraphs or documents, testing joint recall in a single response.

\noindent\textbf{Inferential reasoning.} Questions requiring inference that combines multiple facts from the source document, including temporal reasoning, comparison, and conditional inference. For TimeQA-derived pages, this category includes the original TimeQA questions alongside additional LLM-generated questions.

\noindent\textbf{fail@k} (\textsc{Counter} only) For each single fact recall question, we sample k responses and count an item as a fail if the prior fact appears in any sample, measuring the prior-belief leakage.

Table \ref{tab:examples} shows example document excerpts and questions of each type for both benchmarks.

%% file: body/04_golden_inject_RL.tex
\vspace{-0.5em}

\section{GRIN: Golden-GRPO Injection}
\vspace{-0.5em}

\begin{figure*}[t]
    \centering
    \makebox[\textwidth]{\includegraphics[width=1\textwidth]{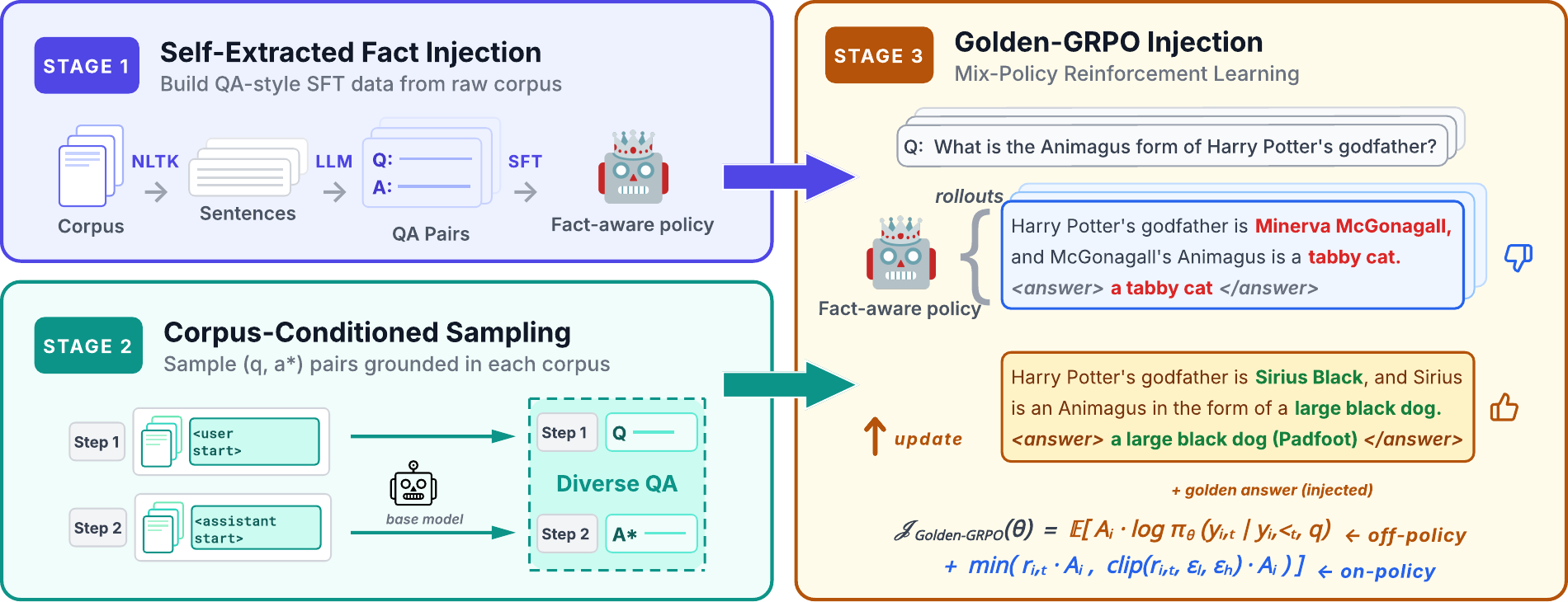}}
    \caption{Overview of the GRIN framework. Stage 1 builds QA-style SFT data from raw corpora; Stage 2 samples diverse (question, golden answer) pairs grounded in each document; Stage 3 trains the model via Golden-GRPO, a mixed-policy RL objective that injects the golden answer as an off-policy trajectory alongside on-policy rollouts.}
    \label{fig:your_label}
\end{figure*}

GRIN is a three-stage framework for continual knowledge injection that operates entirely from the base model, requiring no external teacher (Figure \ref{fig:your_label}). Stages 1 and 2 use the base model to extract atomic QA pairs and to sample a diverse (question, golden-answer) pool from each corpus. Stage 3 trains the model on this pool using Golden-GRPO, our mixed-policy reinforcement learning algorithm that injects the off-policy golden answer to provide learning signal even when on-policy rollouts fail on novel facts, producing knowledge that is absorbed into the model's parametric reasoning rather than memorized in surface form.
\vspace{-0.5em}
\subsection{Stage 1: Self-Extracted Fact Injection}
\vspace{-0.5em}

The first stage extracts atomic facts from source corpora and trains the model to recall them via supervised fine-tuning (SFT). We use only the base model itself for fact extraction, consistent with our self-supervised framing.

Let $D = \{d_1, d_2, d_3, ..., d_N\}$ denote the target corpora, and let $\pi_\theta$ denote the base model with parameter $\theta$. For each corpus $d_i$, we segment it into sentences through the NLTK \cite{bird2006nltk} sentence tokenizer $S$:
\vspace{-0.5em}

\begin{equation}
    S(d_i) = \{s_{i, 1}, s_{i, 2}, s_{i,3}, ..., s_{i, M_i}\}
\end{equation}
For each sentence $s_{i, j}$, we use base model $\pi_\theta$ with fixed prompt $P_{ext}$ to produce atomic factoid question-answer pairs:
\begin{equation}
    F_{i,j} = \pi_\theta(P_{ext} \oplus s_{i, j}) = \{q_{i, j, k}, a_{i, j, k}\}^{K_{i, j}}_{k=1}
\end{equation}
where each $(q,a)$ is a question-answer pair encoding a single fact from the sentence $s_{i, j}$. The full extraction prompt is given in Appendix \ref{app:stage2_example}.

Aggregating across all sentences in all corpora creates a training set $F_{1} = \bigcup_{i,j} F_{i, j}$. We then fine-tune $\pi_\theta$ through standard cross-entropy loss over the answer tokens:
\begin{equation}
    \mathcal{L}_{1}(\theta) = -\mathbb{E}_{(q, a)\sim F_1}\text{log}\pi_\theta(a|q)
\end{equation}

Stage 1 provides the model with parametric access to the injected facts through SFT. However, as shown in Table \ref{tab:main-results}, SFT alone produces brittle memorization. Facts are recalled in their training format but fail on alternative phrasings or compositional queries, which motivates the following stages.
\vspace{-0.5em}

\subsection{Stage 2: Corpus-Conditioned Sampling}
\vspace{-0.5em}

The second stage constructs the (question, golden-answer) pool used as training data for Golden-GRPO in Stage 3. Following prior work that uses pre-query tokens to elicit diverse instruction data from language models \cite{xu2025magpie}, we adapt this technique to a corpus-conditioned setting: for each corpus $d_i \in D$, we prepend corpus and a system prompt before a pre-query token $T_{query}$ to sample questions grounded in the document content, then produce the golden answer a via a pre-answer token $T_{ans}$:
\begin{equation}
    q \sim \pi_\theta(T_{query} \oplus d_{i}),  a^\star \sim \pi_\theta(T_{ans} \oplus d_i \oplus q)
\end{equation}
Each corpus is sampled multiple times with duplicates removed, resulting a pool $F_2 = \{(q_{i,j}, a^*_{i, j})\}$ that varies in phrasing, granularity, and fact coverage. Example outputs are provided in Appendix \ref{app:stage2_example}.
Two design choices distinguish $F_2$ from $F_1$. First, as $F_2$ is used as RL training data rather than SFT targets, we prioritize coverage over individual question quality, that noisy questions that would mislead SFT training may still yield meaningful learning signal through Golden-GRPO reward function. Second, more importantly, $F_2$ is sampled at the corpus level, producing questions that span multiple facts and capture inter-sentence relationships, exactly the patterns that single-sentence extraction cannot generate.

\vspace{-0.5em}

\subsection{Stage 3: Golden-GRPO}


In knowledge injection, on-policy RL on unlearned facts frequently produces zero reward rollouts, yielding zero advantage with no meaningful learning signal. Mixed-policy RL restores non-zero reward to the group by adding an off-policy answer, but importance-weighted off-policy gradients vanish when $\pi_\theta(a^\star)$ is small, leaving only the on-policy disadvantage signal, which drives forgetting without producing recall. We therefore design Golden-GRPO, which replaces the importance-weighted off-policy branch with a direct supervised gradient scaled by the off-policy advantage. This gradient is strong when the model has not yet learned the fact and naturally diminishes as the model improves, smoothly transitioning training toward on-policy exploration. We provide a formal analysis of these dynamics in Appendix \ref{app:dynamics} and empirical evidence in Section \ref{sec:exp}.

\noindent\textbf{Rollout Injection.} For each $(q, a^\star) \in F_2$, we form a rollout group consisting of $N_{on}$ on-policy trajectories sampled from the current policy, together with the off-policy golden answer:
\begin{equation}
    \mathcal{G}(q) = \{\tau_1, \tau_2, ...,\tau_{N_{on}}\} \cup \{\tau^\star\}
\end{equation}
where $\tau_i\sim\pi_{\theta_{old}}(\cdot|q)$ for $i = 1, 2, ..., N_{on}$, and $\tau^\star$ is the trajectory corresponding to the golden answer $a^\star$. Each trajectory receives a scalar reward $R(\tau_i)$.

\noindent\textbf{Training Objective.}
Recall that mixed-policy RL fails in knowledge injection because the gradient toward the golden answer collapses when $\pi_\theta(a^\star)$ is small. Our design protects the gradient at two points.

First, to preserve the absolute magnitude of $A^\star$ in high-variance early-training groups, we adopt Dr.\ GRPO--style group-relative advantages~\cite{liu2025understanding}, which omit the standard-deviation normalization used in GRPO:
\begin{equation}
    A_i = R(\tau_i) - \text{mean}(\{R(\tau_j)\}_{j \in \mathcal{G}(q)})
\end{equation}

Standard GRPO's std normalization shrinks advantages precisely when within-group reward variance is high, which is the exact situation in early knowledge injection training, where a correct off-policy trajectory sits alongside still-failing on-policy rollouts.

We further remove the importance ratio and clip from the off-policy branch and replace them with a direct supervised gradient scaled by the off-policy advantage:
\vspace{-0.5em}
\begin{equation}
\begin{aligned}
     \mathcal{J}&_{\text{Golden-GRPO}}(\theta) =\  \mathbb{E}\bigg[ \underbrace{A_i \cdot \log \pi_\theta(y_{i,t} \mid y_{i,<t}, q)}_{\text{off-policy branch}} \\
    & + \underbrace{\min\Big( r_{i,t} \cdot A_i,\ \text{clip}\big(r_{i,t},\varepsilon_l, \varepsilon_h) \cdot A_i \Big)}_{\text{on-policy branch}} \bigg] 
\end{aligned}
\end{equation}
The off-policy supervised branch contains no importance-sampling ratio, guaranteeing the model is pulled directly toward $a^\star$, scaled by the off-policy advantage $A^\star$. This term dominates early updates when on-policy rollouts fail to produce correct answers. As the model improves and on-policy rollouts begin producing correct answers, $A^\star$ shrinks relative to the on-policy group, and the gradient transitions naturally toward on-policy exploration. A formal analysis of this gradient transition is provided in Appendix \ref{app:dynamics}, with empirical results in Section \ref{sec:exp}.

\noindent\textbf{Reward Design.} The reward function $R(\tau_i)$ combines four components, with distinct roles:

\noindent\textbullet\ \textbf{Format.} Following RL training for mathematical reasoning, we prompt the model to recall relevant facts and then provide the final answer between \texttt{<answer>...</answer>} tags. A correctly-formatted output receives a format reward of $+0.5$.

\noindent\textbullet\ \textbf{ROUGE-L.} With valid format, we extract the content inside the \texttt{<answer>} tags and compute ROUGE-L score against $a^\star$, producing a smooth reward in $[0,0.5]$ based on lexical overlap.

\noindent\textbullet\ \textbf{Exact Match.} To capture exact factual correctness, we extract a target keyword $k^\star$ from $a^\star$ and award $+1$ when the answer contains $k^\star$ as an exact substring.

\noindent\textbullet\ \textbf{Multi-answer penalty.} We observed that the model occasionally emits multiple \texttt{<answer>} blocks when confused. We apply a $-0.25$ penalty for any output containing more than one \texttt{<answer>} block to prevent this reward-hacking pattern.

%% file: body/05_experiment.tex
\begin{table*}[t]
\centering
\caption{Main results on \textsc{Blank} and \textsc{Counter} with Qwen3-4B. Training-free methods (Closed-book, Open-book, RAG) are reference points and are excluded from ranking. \noindent\textbf{Bold} marks the best and \underline{underline} marks the second-best per column within the training-based group; fail@k ($\downarrow$) applies to \textsc{Counter} only.}
\label{tab:main-results}
\footnotesize
\setlength{\tabcolsep}{4pt}
\renewcommand{\arraystretch}{1.08}
\resizebox{\textwidth}{!}{%
\begin{tabular}{l cccc @{\hspace{1.6em}} ccccc}
\toprule
\midrule
& \multicolumn{4}{c}{\noindent\textbf{\textsc{Blank}} (Qwen3-4B)}
& \multicolumn{5}{c}{\noindent\textbf{\textsc{Counter}} (Qwen3-4B)} \\
\cmidrule(lr){2-5} \cmidrule(lr){6-10}
\noindent\textbf{Method}
& \noindent\textbf{Single}\,$\uparrow$ & \noindent\textbf{Multi}\,$\uparrow$ & \noindent\textbf{Infer.}\,$\uparrow$ & \noindent\textbf{Avg}\,$\uparrow$
& \noindent\textbf{Single}\,$\uparrow$ & \noindent\textbf{Multi}\,$\uparrow$ & \noindent\textbf{Infer.}\,$\uparrow$ & \noindent\textbf{Avg}\,$\uparrow$ & \noindent\textbf{fail@k}\,$\downarrow$ \\
\midrule
\multicolumn{10}{l}{\textcolor{gray}{\textit{gemini-3.1-flash-lite}}} \\
Closed-book      & 24.63\% & 5.46\% & 34.43\% & 21.51\% & 1.31\% & 0.00\% & 2.22\% & 1.18\% & 98.53\% \\
Open-book        & 97.07\% & 92.83\% & 95.24\% & 94.05\% & 99.08\% & 100\% & 99.45\% & 99.51\% & 0.00\% \\
\midrule
\multicolumn{10}{l}{\textcolor{gray}{\textit{Training-free}}} \\
Closed-book      & 11.43\% &  1.02\% &  0.55\% &  4.33\% &  2.83\% &  0.00\% &  4.16\% &  2.33\% & 89.92\% \\
Open-book        & 76.54\% & 34.81\% & 36.61\% & 49.32\% & 84.08\% & 43.12\% & 45.43\% & 57.54\% &  2.52\% \\
RAG              & 69.50\% & 11.26\% & 39.34\% & 40.03\% & 70.79\% &  7.34\% & 37.67\% & 38.60\% &  8.93\% \\
\midrule
\multicolumn{10}{l}{\textcolor{gray}{\textit{Training-based}}} \\
PIT              & 28.45\% &  0.68\% &  4.37\% & 11.17\% & 22.79\% &  4.59\% & 20.22\% & 15.87\% & 46.96\% \\
Auton. Learning  & 43.40\% &  2.05\% &  6.56\% & 17.34\% & 23.83\% &  1.83\% & 21.05\% & 15.57\% & 37.57\% \\
Self-Tuning      & \noindent\textbf{54.54\%} & \underline{11.43\%} &  7.10\% & \underline{24.30\%} & 35.05\% &  8.26\% & \underline{30.47\%} & 24.59\% & 38.14\% \\
\cmidrule(lr){1-10}
\multicolumn{10}{l}{\textcolor{gray}{\textit{Ablations of \textsc{Grin}}}} \\
~~\textsc{Grin} w/ SFT   & 27.57\% &  5.46\% &  9.29\% & 14.11\% & \underline{36.43\%} &  0.92\% & 20.78\% & 19.38\% & 39.63\% \\
~~\textsc{Grin} w/ GRPO & 5.28\% &  0.00\% & 0.21\% & 1.83\% & 23.02\% & 3.67\% & 15.79\% & 14.16\% & \underline{24.86\%} \\
~~\textsc{Grin} w/ LUFFY & 35.19\% &  9.56\% & \underline{16.39\%} & 20.38\% & 34.71\% & \underline{25.69\%} & 28.53\% & \underline{29.64\%} & 31.96\% \\
\textsc{Grin} (ours)     & \underline{52.49\%} & \noindent\textbf{21.16\%} & \noindent\textbf{31.69\%} & \noindent\textbf{35.11\%} & \noindent\textbf{54.64\%} & \noindent\textbf{34.37\%} & \noindent\textbf{42.38\%} & \noindent\textbf{43.65\%} & \noindent\textbf{24.05\%} \\
\bottomrule
\end{tabular}%
}
\end{table*}

\begin{table}[t]\label{table:llama}
\centering
\caption{Cross-model results on COUNTER with Llama3.2-3B. We evaluate on COUNTER as counter facts are model irrelevant. GRIN leads on three of four accuracy metrics and is competitive on fail@k, indicating that the contribution generalizes beyond Qwen3-4B.}
\label{tab:llama3b}
\footnotesize
\setlength{\tabcolsep}{3.5pt}
\renewcommand{\arraystretch}{1.08}
\resizebox{\columnwidth}{!}{%
\begin{tabular}{l ccccc}
\toprule
\midrule
\multicolumn{6}{c}{\textit{\noindent\textbf{\textsc{Counter}}}(Llama3.2-3B)} \\
\midrule
\noindent\textbf{Method} &
\noindent\textbf{Single}\,$\uparrow$ &
\noindent\textbf{Multi}\,$\uparrow$ &
\noindent\textbf{Infer.}\,$\uparrow$ &
\noindent\textbf{Avg}\,$\uparrow$ &
\noindent\textbf{fail@k}\,$\downarrow$ \\
\midrule
PIT                 & 29.10\% & \underline{9.17\%} & 27.98\% & 22.08\% & 38.60\% \\
AL (DPO)            & 29.67\% & 2.75\% & 24.10\% & 18.84\% & 35.97\% \\
Self-Tuning         & \underline{39.63\%} & 8.26\% & \underline{33.80\%} & \underline{27.23\%} & \noindent\textbf{27.38\%} \\
\textsc{Grin} (ours)& \noindent\textbf{42.38\%} & \noindent\textbf{11.01\%} & \noindent\textbf{37.95\%} & \noindent\textbf{30.45\%} & \underline{27.72\%} \\
\bottomrule
\end{tabular}%
}
\end{table}

\vspace{-0.5em}
\section{Experiment}\label{sec:exp}

\subsection{Experimental Setup}

\noindent\textbf{Models and Infrastructure.} We use Qwen3-4B\cite{yang2025qwen3} as the base model for all main results. All experiments are run on 2×H200 GPUs. We evaluate on \textsc{Blank} and \textsc{Counter} (Section \ref{sec:benchmark}), reporting per-type accuracy and the per-benchmark average; for \textsc{Counter} we additionally report fail@k with k=5. Accuracy is judged by an LLM (gemini-3.5-flash), with prompt details provided in Appendix \ref{app:judge_prompts}.

\noindent\textbf{Baselines.} We compare GRIN against both training-free and training-based methods to fully evaluate how mixed-policy reinforcement learning contributes to continual knowledge injection. For training-free baselines, \textit{closed-book} queries the base model without context and serves as a sanity check, that near-zero accuracy confirms our evaluation questions cannot be answered from the model's current knowledge base alone. \textit{Open-book} prompts the base model with the source wiki page in the context and serves as an upper-bound reference for knowledge injection. \textit{RAG\cite{lewis2020retrieval}} retrieves at inference time with top-k set to 3. For training-based methods, \textit{PIT}\cite{jiang2024instruction} trains with QA pairs positioned before their corresponding document texts. \textit{Self-tuning}\cite{zhang2025self} extends PIT with self-generated diverse training data, and \textit{Autonomous Learning (AL)} \cite{ji2025unlocking} extends supervised training with offline direct preference optimization (DPO) \cite{rafailov2023direct}.

\noindent\textbf{Implementation.} We use the AdamW \cite{loshchilov2017decoupled} optimizer with learning rate 2e-5 for both Stage 1 SFT and Stage 3 Golden-GRPO training, with an effective batch size of 512. Stage 1 is trained for 5 epochs and Stage 3 Golden-GRPO for 3 epochs with 8 on-policy rollouts. We set the KL coefficient to 5 which is substantially higher than the 0.01–0.1 range typical in RLHF, as we empirically observe knowledge injection induces large parameter shifts in early training.

\subsection{Main Results}
\vspace{-0.5em}


\noindent\textbf{Benchmark Validation.} The training-free references in Table \ref{tab:main-results} show that the benchmarks are well-designed. Closed-book accuracy is near zero on both \textsc{Blank} and \textsc{Counter}, with averages of 4.33\% and 2.33\% respectively, confirming that the evaluation questions cannot be answered from the base model's parametric knowledge alone. Open-book achieves high accuracy on single-fact recall but only 36.61\% and 45.43\% on inferential reasoning, indicating that the inferential questions are genuinely difficult. RAG performs strongly on single-fact recall but collapses on multi-source retrieval as our multi-source questions span facts across pages that retrieval cannot reliably surface. Additionally, a close-sourced model (gemini-3.1-flash-lite) with documents in context reaches 94.05\% on \textsc{Blank} and 99.51\% on \textsc{Counter}, confirming our benchmark questions are answerable from the corpora.

\noindent\textbf{SFT memorize but does not generalize.} Among training based baselines, PIT, AL, and Self-Tuning all have respectable single-fact recall on both \textsc{Blank} and \textsc{Counter}, but degrade sharply on multi-source retrieval and inferential reasoning tasks. This is the characteristic failure mode of supervised injection, that facts are bound to their training questions forms but cannot be retrieved under different phrasings or question types. This can be clearly observed on Self-Tuning, the strongest baseline, which achieves $54.54\%$ single fact recall in \textsc{Blank}, but drops to only $7.10\%$ on inferential reasoning, an almost 8 times drop across question types that rely on the same underlying knowledge.

\noindent\textbf{Golden-GRPO enables absorption.} GRIN is competitive with the strongest SFT-based baseline on single-fact recall, confirming that the RL objective does not sacrifice basic recall. On the more generalized question types, the gap is more significant. GRIN reaches 21.16\% multi-source and 31.69\% inferential on \textsc{Blank}, compared to Self-Tuning's 11.26\% and 7.10\%. The \textsc{Counter} results show the same pattern, and GRIN's fail@k of 24.05\% is lower than every other training-based method, indicating that GRIN can not only learn new facts, but suppresses prior beliefs as well. 

\noindent\textbf{Golden-GRPO's design is necessary.} We further shown in ablation that the performance gain is not the result of diverse sampled data alone. Continue training on the same Stage 2 data on Stage 1 model with SFT yields similar performance to other SFT baselines on multi-source retrieval and inferential reasoning, empirically proving that the training method, not the data, is the key obstacle to generalize the injected knowledge. On-policy RL alone is even less effective, that vanilla GRPO without off-policy injection collapses performance to near zero on BLANK (1.83\% average), confirming that without the off-policy golden trajectory, on-policy rollouts produce no learning signal on facts the model has not yet acquired. Moreover, existing mixed-policy RL method is not suffice as well. Replacing Golden-GRPO with LUFFY \cite{yan2026learning} outperforms SFT on generalization but underperforms ours, indicating that mixed-policy RL is a viable direction and the design choices in Golden-GRPO closed the remaining gap.

\noindent\textbf{Cross-model generalization.} To verify that GRIN's gains are not specific to Qwen3-4B, we replicate the \textsc{Counter} experiment with Llama3.2-3B \cite{grattafiori2024llama} as the base model. \textsc{Counter} replacements contradict real-world facts, so for any base model they serve as valid injection targets, that either overwriting existing beliefs or acquiring novel ones. Table \ref{tab:llama3b} reports results: GRIN leads on three of four accuracy metrics and is competitive on fail@k, reproducing the memorization-versus-absorption pattern observed with Qwen3-4B and indicating that the contribution generalizes across base models.

\begin{figure}[t]\label{fig:long}
    \centering
    \includegraphics[width=\columnwidth]{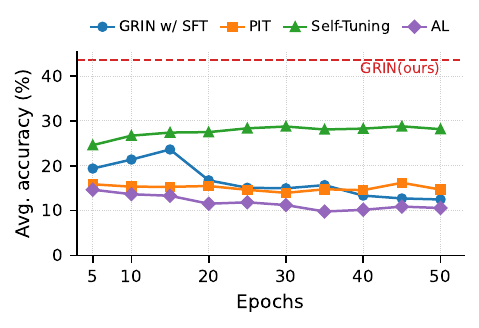}
    \caption{Average \textsc{Counter} accuracy across training epochs. SFT-based baselines plateau or degrade well below GRIN (dashed) even at 50 epochs, showing the generalization gap is not closed by additional compute.}
    \vspace{-0.8em}
    \label{fig:long-training-curves}
\end{figure}

\noindent\textbf{Compute-matched training.} A natural question is whether GRIN's advantage stems simply from its higher training compute relative to SFT based baselines. To test this, we continual train all baselines for additional epochs, resulting in a total of 50 epochs, which match or even exceed the total training cost of Golden-GRPO. Figure \ref{fig:long-training-curves} reports \textsc{Counter} average accuracy across training epochs. Training-based baselines plateau beyond 15 epochs, well below GRIN's 43.65\% average accuracy. This indicates that the gap is not closed by additional training. SFT based methods reach a ceiling determined by their training objective, while GRIN's RL objective is capable of extract gains from the same data, unlocking performance that SFT cannot reach regardless of compute.

%% file: body/06_conclusion.tex
\vspace{-0.5em}
\section{Conclusion}
\vspace{-0.5em}
In this paper, we presented GRIN, a three-stage framework for continual knowledge injection built around Golden-GRPO, a mixed policy reinforcement learning algorithm tailored to this setting. Motivated by the failure of supervised methods to absorb knowledge beyond their training format, and by the vanishing off-policy gradient that limits standard mixed-policy RL on novel facts, Golden-GRPO produces knowledge that is absorbed into the model's parametric reasoning rather than memorized in surface form. Across novel acquisition (\textsc{Blank}) and counterfactual overwrite (\textsc{Counter}) benchmarks, GRIN substantially outperforms supervised and mixed-policy RL baselines on multi-source retrieval and inferential reasoning. 


%% file: body/07_limitations.tex
\section*{Limitations}
While our experimental results have demonstrated that mixed-policy reinforcement learning enables knowledge absorption beyond what SFT achieves, several aspects of this work remain open for further investigation.

\noindent\textbf{Single-round injection vs. Lifelong learning.} Our benchmarks evaluate a single round of knowledge injection, where the model learns from a group of corpora. However, the goal of lifelong learning is continual training on the target model as the real world evolves. A successful single-round injection does not guarantee successful multi-round injection. Each round of training shapes the model's parameter space in ways that may affect both the retention of previously injected knowledge and the model's capacity to absorb future knowledge. How Golden-GRPO behaves under repeated injection rounds, whether it accumulates knowledge stably, suffers from catastrophic forgetting, or degrades the model's ability to learn remains an open question, and require further investigation.

\noindent\textbf{Knowledge domain coverage.} Our benchmarks (\textsc{Blank} and \textsc{Counter}) are constructed from wiki-style, entity-grounded text. The behavior of GRIN on other knowledge formats, like procedural knowledge, code, mathematical content, or structured data, is unexplored. The reward function (ROUGE-L and Exact Match) is tailored to factual recall and may need adaptation for domains where correctness is harder to measure with lexical signals.

\noindent\textbf{Resource Requirements.}
Golden-GRPO requires sampling on-policy rollouts at each training step, which scales with model size, dataset size and rollout count. At the 3–4B scale we evaluate, this overhead is modest and is justified by our compute-matched experiments. At frontier scales, however, the cost of rollout sampling on a trillion-parameter model with world-scale knowledge updates may approach the cost of full retraining. In such cases, full retraining remains preferable because it offers more than knowledge updates, as it also allows architectural improvements (e.g. new attention mechanism \cite{team2026attention}) that knowledge injection cannot provide. Our method is therefore most applicable to scenarios where retraining the base model is not an option, like task-specific customization, company-internal knowledge updates, or rapid deployment of new factual content. As foundation model architectures mature and retraining shifts toward serving primarily for parametric knowledge updates, Golden-GRPO's specialized approach to knowledge injection may become attractive at larger scales as well.

%% file: body/08_ethic.tex
\section*{Ethical considerations}
The \textsc{Counter} benchmark contains counterfactual statements that contradict real-world facts; these are constructed strictly as evaluation infrastructure and are not intended for deployment.

%% file: body/appendix.tex
\section{Prompts used for corpus construction}\label{app:counterprompts}

We construct the \textsc{Counter} corpora using two API calls per subject.
Prompt~1 (Table~\ref{tab:prompt-prior}) elicits the base model's prior
knowledge of a subject as a short Wikipedia-style passage. Prompt~2
(Table~\ref{tab:prompt-counter}) takes that passage and rewrites it into a
parallel-universe version in which only identifiable concrete facts are
replaced, while structure and the subject entity are preserved.

\begin{table*}[]
\centering
\small
\begin{tabularx}{\textwidth}{@{}lX@{}}
\toprule
\textbf{Component} & \textbf{Content} \\
\midrule
System prompt &
You are a knowledgeable encyclopedia writer.
\\
\addlinespace
User template &
Write a concise, factual Wikipedia-style article about the
\texttt{\{entity\_type\}}, \texttt{\{entity\_name\}}.

Keep the article informative and tightly written, covering only the most
essential facts, dates, and notable details. Organize the text using
exactly 3 to 4 standard markdown headings (\#\#) that are most appropriate
for this specific subject. Each section should be a focused, well-developed
paragraph --- avoid bullet points, repetition, and unnecessary elaboration.
Write in an encyclopedic, neutral tone. \\
\bottomrule
\end{tabularx}
\caption{Prompt 1: sampling the base model's current beliefs about a subject.
}
\label{tab:prompt-prior}
\end{table*}

\begin{table*}[]
\centering
\small
\begin{tabularx}{\textwidth}{@{}lX@{}}
\toprule
\textbf{Component} & \textbf{Content} \\
\midrule
User template &
I want you to design a parallel-universe version of the provided wiki page.

Your task is to replace concrete facts in the text --- proper nouns,
dates, years, numerical values, locations, named entities, organizations,
award names, and titles --- with plausible, internally consistent
alternatives that could believably appear in a real wiki. Do \textbf{not}
alter common nouns (e.g.\ ``novel'', ``painter''), generic descriptors, or
linguistic connective tissue --- only swap identifiable concrete facts.

\smallskip
\textbf{Strict rules:}
\begin{enumerate}\itemsep0pt\parsep0pt
\item Do \textbf{not} change the main entity/subject of the wiki page,
      which is: \texttt{\{subject\}}.
\item Preserve section headers verbatim, preserve paragraph count and
      order, and keep sentences in the same positions. Only the concrete
      facts inside sentences change.
\item If a name, date, place, or other replaced entity appears multiple
      times in the original, the \textbf{same} substitute must be used
      at every occurrence.
\item Output \textbf{only} the rewritten article in markdown --- no
      preamble, no closing commentary, no annotations or markers in the
      prose.
\end{enumerate}

\smallskip
Here is the original wiki page content to transform:

\texttt{\{content\}} \\
\bottomrule
\end{tabularx}
\caption{Prompt 2: rewriting a passage from Prompt~1 into a
``parallel-universe'' version.}
\label{tab:prompt-counter}
\end{table*}

\section{Prompts used for question generation}\label{app:questionprompts}
We construct the question pool for \textsc{Blank} and \textsc{Counter} using
three LLM prompts, one per question type. Prompt~1 (Table~\ref{tab:prompt-single})
generates single-fact recall questions from a single sentence of the source
document. Prompt~2 (Table~\ref{tab:prompt-multi}) generates multi-source
retrieval questions composed of $N \in \{2, 3, 4\}$ sub-questions targeting
distinct facts that span different paragraphs or documents. Prompt~3
(Table~\ref{tab:prompt-infer}) generates inferential reasoning questions that
require combining multiple facts from the source document. For \textsc{Blank}'s
inferential reasoning category, we additionally include the original TimeQA
questions for TimeQA-derived passages; Prompt~3 is used only for the
LLM-augmented portion and for all post-cutoff and \textsc{Counter} passages.

\begin{table*}[]
\centering
\small
\begin{tabularx}{\textwidth}{@{}lX@{}}
\toprule
\textbf{Component} & \textbf{Content} \\
\midrule
System prompt &
You are an expert question writer for reading-comprehension evaluation.
Your job is to produce factoid questions whose answers are unambiguous
and directly supported by a single sentence in the source document. \\
\addlinespace
User template &
Read the following document and generate a set of single-fact recall
questions. Each question must satisfy all of the following:
\begin{enumerate}\itemsep0pt\parsep0pt
\item The answer is a short concrete fact (a name, date, place, number,
      or named entity) that appears verbatim in exactly one sentence of
      the document.
\item The question is fully answerable from the document without
      additional context or outside knowledge.
\item The question form does not paraphrase the source sentence;
      rephrase it so that a model that has merely memorized the source
      cannot trivially pattern-match.
\item Each question targets a different fact; do not generate
      paraphrases of the same question.
\end{enumerate}
\smallskip
For each question, output a JSON object with fields
\texttt{question}, \texttt{answer}, and \texttt{source\_sentence}.
Output one JSON object per line, no additional commentary.
\smallskip
Document:
\texttt{\{document\}} \\
\bottomrule
\end{tabularx}
\caption{Prompt for generating single-fact recall questions.}
\label{tab:prompt-single}
\end{table*}

\begin{table*}[]
\centering
\small
\begin{tabularx}{\textwidth}{@{}lX@{}}
\toprule
\textbf{Component} & \textbf{Content} \\
\midrule
System prompt &
You are an expert question writer who specializes in composite
questions that test joint recall across multiple facts. \\
\addlinespace
User template &
Read the following document(s) and generate multi-source retrieval
questions. Each question must be composed of \texttt{\{N\}}
sub-questions ($N \in \{2, 3, 4\}$), where each sub-question targets a
distinct concrete fact, and the sub-questions reference facts that
appear in different paragraphs or different documents.
\smallskip
Requirements:
\begin{enumerate}\itemsep0pt\parsep0pt
\item Sub-question facts must not overlap; each targets a different
      entity, date, location, or attribute.
\item The composite question must be phrased as a single natural
      question (e.g.\ ``What is X and when was Y?'') rather than a list.
\item The golden answer must list each sub-answer in the order the
      sub-questions appear, separated by semicolons.
\item Do not generate questions whose sub-facts could all be retrieved
      from a single paragraph; the point is to test joint recall across
      separated passages.
\end{enumerate}
\smallskip
For each question, output a JSON object with fields
\texttt{question}, \texttt{answer}, \texttt{N}, and
\texttt{source\_locations} (list of paragraph or document identifiers
for each sub-fact). Output one JSON object per line, no additional
commentary.
\smallskip
Document(s):
\texttt{\{documents\}} \\
\bottomrule
\end{tabularx}
\caption{Prompt for generating multi-source retrieval questions.}
\label{tab:prompt-multi}
\end{table*}

\begin{table*}[]
\centering
\small
\begin{tabularx}{\textwidth}{@{}lX@{}}
\toprule
\textbf{Component} & \textbf{Content} \\
\midrule
System prompt &
You are an expert question writer who designs inference questions that
require combining multiple facts from a source document. \\
\addlinespace
User template &
Read the following document and generate inferential reasoning
questions. Each question must satisfy all of the following:
\begin{enumerate}\itemsep0pt\parsep0pt
\item Answering the question requires combining at least two distinct
      facts from the document. Direct retrieval of any single fact
      should be insufficient.
\item The inference type should fall into one of: temporal reasoning
      (e.g.\ computing durations, ordering events), comparison
      (e.g.\ identifying which of several entities has a property),
      or conditional inference (e.g.\ ``given X, what does Y imply'').
\item The golden answer must be a short concrete value (a number, date,
      name, or short phrase). Do not generate questions whose answers
      are open-ended explanations.
\item The required inference must be derivable from the document alone,
      without external world knowledge.
\end{enumerate}
\smallskip
For each question, output a JSON object with fields
\texttt{question}, \texttt{answer}, \texttt{inference\_type}
(\texttt{temporal} / \texttt{comparison} / \texttt{conditional}), and
\texttt{supporting\_facts} (list of source sentences used). Output one
JSON object per line, no additional commentary.
\smallskip
Document:
\texttt{\{document\}} \\
\bottomrule
\end{tabularx}
\caption{Prompt for generating inferential reasoning questions.}
\label{tab:prompt-infer}
\end{table*}

\section{Example Stage 2 QA pair for Golden-GRPO}\label{app:stage2_example}

To illustrate Stage 2's corpus-conditioned sampling mechanism, we show
how the pre-query token elicits a diverse question from the base model
conditioned on a source corpus. Following~\citet{xu2025magpie}, we
construct a prompt by prepending the corpus and a system prompt before
the pre-query token, and let the base model continue from there. The
model's continuation is then truncated at the first end-of-turn marker
to extract the sampled question. Table~\ref{tab:stage2-query} shows a
representative example using a \textsc{Counter} corpus on Apple's stock
listing (where the real-world fact is Apple on NASDAQ under
\texttt{AAPL}, but the corpus has been rewritten to place Apple on the
NYSE under \texttt{APL}).

\begin{table*}[]
\centering
\small
\begin{tabularx}{\textwidth}{@{}lX@{}}
\toprule
\textbf{Field} & \textbf{Content} \\
\midrule
System prompt &
You are a helpful AI assistant that answers questions about provided
documents. \\
\addlinespace
Corpus &
\#\# Listing and Trading
\smallskip
Apple Inc.\ is listed on the New York Stock Exchange (NYSE) under the
ticker symbol \texttt{APL}....
\smallskip
[\textit{remainder of corpus elided for space}] \\
\addlinespace
Pre-query token &
\texttt{<|im\_start|>user} \\
\addlinespace
Sampled question (model continuation) &
What is the ticker symbol for Apple on the NYSE? \\
\bottomrule
\end{tabularx}
\caption{Stage 2 question sampling using a pre-query token.}
\label{tab:stage2-query}
\end{table*}

\begin{table*}[]
\centering
\small
\begin{tabularx}{\textwidth}{@{}lX@{}}
\toprule
\textbf{Field} & \textbf{Content} \\
\midrule
Question &
What is the ticker symbol for Apple on the NYSE? \\
\addlinespace
Golden answer &
Apple is a major publicly traded corporation that maintains a
significant presence on the New York Stock Exchange. As part of its
financial operations and public listing, the company is identified by
a specific ticker symbol used for trading its shares. That ticker
symbol is APL.
\smallskip
\texttt{<answer>The ticker symbol for Apple on the NYSE is APL.</answer>} \\
\bottomrule
\end{tabularx}
\caption{A representative (question, golden-answer) data pair from Stage 2
sampling on \textsc{Counter}.}
\label{tab:stage2-example}
\end{table*}

\section{Judge Prompts}\label{app:judge_prompts}
We use two LLM-as-judge prompts for evaluation. The accuracy judge
(Table~\ref{tab:prompt-accuracy}) determines whether a model's answer
contains the same factual information as the reference answer; this is
used for all per-type accuracy scores reported in Tables 2 and 3. The
fail@k judge (Table~\ref{tab:prompt-failk}) is used only for
\textsc{Counter}, classifying each of $k$ sampled answers as agreeing
with the counterfactual training text, the original Wikipedia prior, or
neither.

\begin{table*}[t]
\centering
\small
\begin{tabular}{@{}lp{0.82\textwidth}@{}}
\toprule
\textbf{Component} & \textbf{Content} \\
\midrule
System prompt &
You are a strict factual judge. Your job is to determine whether the
student's answer contains the same factual information as the reference
answer.

Follow these steps:

(1) Extract the core factual claim from the reference answer.

(2) Check if the student's answer contains this same fact.

(3) If the student refuses to answer, says the question is invalid,
says ``no such person/place exists'', or hedges without providing a
concrete answer, that is \texttt{WRONG}.

(4) Ignore any formatting, emojis, self-praise (``correct'',
``accurate''), or repeated statements in the student's answer. Focus
\textbf{only} on the factual content.

(5) Output \textbf{only} the word \texttt{CORRECT} or \texttt{WRONG}. \\
\addlinespace
User template &
Question: \texttt{\{question\}}

Reference Answer: \texttt{\{golden\_answer\}}

Student's Answer: \texttt{\{generated\_answer\}}

Does the student's answer contain the same factual information as the
reference answer? Output \textbf{only}: \texttt{CORRECT} or \texttt{WRONG}. \\
\bottomrule
\end{tabular}
\caption{Prompt used by the accuracy judge to score per-type accuracy.}
\label{tab:prompt-accuracy}
\end{table*}

\begin{table*}[]
\centering
\small
\begin{tabular}{@{}lp{0.82\textwidth}@{}}
\toprule
\textbf{Component} & \textbf{Content} \\
\midrule
User template &
You are judging whether a model's answer reflects the
\textbf{counterfactual} knowledge that the model was trained on, or the
\textbf{original} prior knowledge from real-world Wikipedia.

Question: \texttt{\{question\}}

Model's answer: \texttt{\{model\_answer\}}

\texttt{--- Counterfactual text (the model's training target) ---}

\texttt{\{counterfactual\}}

\texttt{--- end counterfactual ---}

\texttt{--- Original Wikipedia text (real-world prior) ---}

\texttt{\{original\}}

\texttt{--- end original ---}

\textbf{Verdict --- choose ONE:}
\begin{itemize}\itemsep0pt\parsep0pt
\item \texttt{COUNTERFACTUAL} --- the model's answer agrees with the counterfactual text.
\item \texttt{ORIGINAL} --- the model's answer agrees with the original Wikipedia (model fell back to prior).
\item \texttt{NEITHER} --- the answer matches neither (refusal, hallucination, off-topic).
\end{itemize}

Output exactly one word: \texttt{COUNTERFACTUAL}, \texttt{ORIGINAL}, or \texttt{NEITHER}. \\
\bottomrule
\end{tabular}
\caption{Prompt used by the fail@k judge.}
\label{tab:prompt-failk}
\end{table*}

\section{Gradient dynamics of Golden-GRPO vs. importance-weighted mixed-policy RL}\label{app:dynamics}

This appendix formalizes the gradient-vanishing problem of
importance-weighted mixed-policy RL in the knowledge injection setting,
and shows why Golden-GRPO's direct supervised gradient design avoids it.
We use LUFFY~\citep{yan2026learning} and
GOLF~\cite{huang2026bootstrapping} as two representative
importance-weighted mixed-policy baselines that share the same
structural failure mode.

\paragraph{LUFFY's mixed-policy objective.}
LUFFY combines on-policy rollouts with off-policy trajectories drawn from
a reference policy $\pi_\phi$ (in our case, the LLM that generated the
golden answer):
\begin{equation*}
\begin{aligned}
\mathcal{J}_{\text{LUFFY}}(\theta) = \frac{1}{Z}\bigg( & \underbrace{\sum_{j=1}^{N_{\text{off}}} \sum_{t=1}^{|\tau_j|} \text{CLIP}\big(\hat{r}_{j,t}^{\text{LUFFY}},\hat{A}_j, \varepsilon\big)}_{\text{off-policy branch}} \\
& + \underbrace{\sum_{i=1}^{N_{\text{on}}} \sum_{t=1}^{|\tau_i|} \text{CLIP}\big(r_{i,t}(\theta),\hat{A}_i, \varepsilon\big)}_{\text{on-policy branch}} \bigg)
\end{aligned}
\end{equation*}
where the off-policy importance ratio is
$\hat{r}_{j,t}^{\text{LUFFY}} = \pi_\theta(\tau_{j,t} \mid q, \tau_{j,<t}) / \pi_\phi(\tau_{j,t} \mid q, \tau_{j,<t})$.

\paragraph{GOLF's mixed-policy objective.}
GOLF takes a similar form but constructs the off-policy ratio differently.
Rather than drawing trajectories from an external reference policy, GOLF
augments the original prompt $x$ with natural-language feedback or
guidance $p_{\text{agg}}(x)$ that helps the model produce a correct
answer, and treats the resulting trajectory as off-policy data. The
off-policy ratio is:
\begin{equation*}
r_{j,t}^{\text{GOLF}}(\theta) = \frac{\pi_\theta(\tau_{j,t} \mid x, \tau_{j,<t})}{\pi_{\theta_{\text{old}}}(\tau_{j,t} \mid p_{\text{agg}}(x), \tau_{j,<t})}
\end{equation*}
Here the numerator and denominator condition on \emph{different prompts}:
the numerator on the plain query $x$, and the denominator on the
augmented query $p_{\text{agg}}(x)$ that explicitly includes guidance.

\paragraph{Why both ratios vanish in knowledge injection.}
Both LUFFY and GOLF suffer from the same structural problem: the
denominator's probability for the off-policy trajectory $\tau^\star$ is
substantially larger than the numerator's, making the ratio small
precisely when the model has not yet learned the injected fact.

For LUFFY, the reference policy $\pi_\phi$ assigns high probability to
its own generated trajectory, so the denominator $\pi_\phi(\tau^\star)$
is large. The numerator $\pi_\theta(\tau^\star)$ is small in early
training because the model has not learned the fact. The ratio
$\hat{r}^{\text{LUFFY}} = \pi_\theta / \pi_\phi$ collapses.

For GOLF, the asymmetry is even more pronounced. The augmented prompt
$p_{\text{agg}}(x)$ is \emph{explicitly designed} to make $\tau^\star$
likely, that it contains feedback or guidance that steers the model
toward the correct answer. Thus
$\pi_{\theta_{\text{old}}}(\tau^\star \mid p_{\text{agg}}(x))$ is by
construction high, while $\pi_\theta(\tau^\star \mid x)$ on the plain
prompt remains small until the model has actually learned the fact. The
ratio $r^{\text{GOLF}} = \pi_\theta(\cdot \mid x) / \pi_{\theta_{\text{old}}}(\cdot \mid p_{\text{agg}}(x))$
collapses for the same reason as LUFFY.

In both cases, the gradient contribution of the off-policy branch is
proportional to this small ratio, regardless of how large the off-policy
advantage $\hat{A}^\star$ is.

\paragraph{Golden-GRPO's design.}
Golden-GRPO removes the importance ratio entirely from the off-policy
branch and replaces it with a direct supervised gradient scaled by the
off-policy advantage:
\begin{equation*}
\mathcal{J}_{\text{off}}^{\text{Golden}} = A^\star \cdot \log \pi_\theta(y_{i,t} \mid y_{i,<t}, q)
\end{equation*}
The gradient with respect to $\theta$ is:
\begin{equation*}
\nabla_\theta \mathcal{J}_{\text{off}}^{\text{Golden}} = A^\star \cdot \nabla_\theta \log \pi_\theta(y_{i,t} \mid y_{i,<t}, q)
\end{equation*}
The magnitude depends only on the advantage $A^\star$ and the standard
supervised gradient $\nabla_\theta \log \pi_\theta$. Critically, this
does \emph{not} vanish when $\pi_\theta(\tau^\star \mid q)$ is small;
the supervised gradient is well-defined and stable across the range of
$\pi_\theta$ values encountered during training.

\paragraph{Three-phase gradient analysis.}\label{app:gradient}
To make the self-regulating behavior concrete, we trace the gradient
composition through three representative training phases. For
simplicity, assume the reward is binary: $R(\tau) = R_{\max}$ if the
trajectory is correct, $0$ otherwise. Golden-GRPO uses $N_{\text{on}}$
on-policy rollouts plus one off-policy golden trajectory $\tau^\star$
with $R(\tau^\star) = R_{\max}$.

\textit{Phase 1: All on-policy rollouts incorrect.} The group reward
distribution is $\{R_{\max}, 0, 0, \ldots, 0\}$, giving group mean
$\bar{R} = R_{\max} / (N_{\text{on}} + 1)$. The advantages are:
\begin{equation*}
A^\star = R_{\max} \cdot \frac{N_{\text{on}}}{N_{\text{on}}+1}, \quad
A_i = -\frac{R_{\max}}{N_{\text{on}}+1}
\end{equation*}
The off-policy advantage $A^\star$ is large and positive, while
on-policy advantages are small and negative. The gradient is dominated
by the off-policy supervised term, pulling the model strongly toward
$\tau^\star$.

\textit{Phase 2: Mixed rollouts.} Suppose $k$ of $N_{\text{on}}$
on-policy rollouts are correct, the rest incorrect. The group mean
rises to $\bar{R} = (k+1) R_{\max} / (N_{\text{on}}+1)$, and the
advantages become:
\begin{equation*}
A^\star = A_{\text{correct}} = R_{\max} \cdot \frac{N_{\text{on}} - k}{N_{\text{on}}+1}
\end{equation*}
\begin{equation*}
    A_{\text{wrong}} = -R_{\max} \cdot \frac{k+1}{N_{\text{on}}+1}
\end{equation*}
The off-policy advantage has shrunk; correct on-policy rollouts now
share the same positive advantage as the golden trajectory, so they
contribute positively to the policy update. The model reinforces both
the golden trajectory and its own correct outputs simultaneously, and
the off-policy branch's relative dominance over the total gradient
diminishes as $k$ grows.

\textit{Phase 3: All on-policy rollouts correct.} The group reward
distribution becomes uniformly $R_{\max}$, giving $\bar{R} = R_{\max}$
and $A^\star = A_i = 0$ for all trajectories. The entire objective
gradient vanishes for this question. The model has converged on this
fact, and further training resources are naturally redirected to
questions where the rollout group still shows advantage variance.

\paragraph{Why this matters.}
This three-phase progression demonstrates Golden-GRPO's self-regulating
property in concrete form: the off-policy gradient is largest when the
model needs it most (Phase 1), shrinks gracefully as the model learns
(Phase 2), and reaches zero exactly when the model has internalized the
fact (Phase 3). The transition emerges automatically from the advantage
formula and does not require any external schedule or annealing.
Importantly, the model is never \emph{over-pulled} toward the golden
trajectory: once on-policy rollouts succeed at the same rate as the
off-policy reference, the supervised gradient toward $\tau^\star$
disappears, preventing the model from overfitting onto the specific
phrasing of the LLM-generated golden answer.

\section{Reproducibility Statement.}
We use Qwen3-4B\cite{yang2025qwen3}, Llama3.2-3B\cite{grattafiori2024llama}, TimeQA\cite{chen2021dataset}, and NLTK\cite{bird2006nltk} under their respective open-source licenses. Code and benchmark data will be made publicly available; a code and data archive accompanies this submission for reviewer access.

\section{The Use of Large Language Models (LLMs)}
Large language models were used solely to aid in polishing the writing of this paper. They were not used for research ideation, methodology, analysis, or concluding. The authors take full responsibility for all content.